\documentclass[10pt,twocolumn,letterpaper]{article}

\usepackage{iccv}
\usepackage{graphicx}
\usepackage{booktabs}
\usepackage{amsmath}
\usepackage{amsthm}
\usepackage{booktabs}
\usepackage{algorithm}
\usepackage{algorithmic}
\usepackage{epsfig}
\usepackage{amssymb}
\usepackage{bbold}
\usepackage{xcolor}
\usepackage{caption}
\usepackage{multirow}
\usepackage{soul}
\usepackage{bbm}
\usepackage{wrapfig}
\usepackage{comment}
\usepackage{enumitem}
\usepackage{arydshln}
\usepackage[title]{appendix}

\usepackage[pagebackref=true,breaklinks=true,letterpaper=true,colorlinks,bookmarks=false]{hyperref}

\iccvfinalcopy

\ificcvfinal\fi

\begin{document}

\title{Training-free Color-Style Disentanglement for Constrained\\ Text-to-Image Synthesis}

\author{Aishwarya Agarwal, Srikrishna Karanam, and Balaji Vasan Srinivasan\\
Adobe Research, Bengaluru India \\
{\tt \scalebox{.7}{\{aishagar,skaranam,balsrini\}@adobe.com}}
}

\twocolumn[{
\renewcommand\twocolumn[1][]{#1}%
\maketitle
\begin{center}
 \centering
 \captionsetup{type=figure}
 \includegraphics[width=0.7\textwidth]{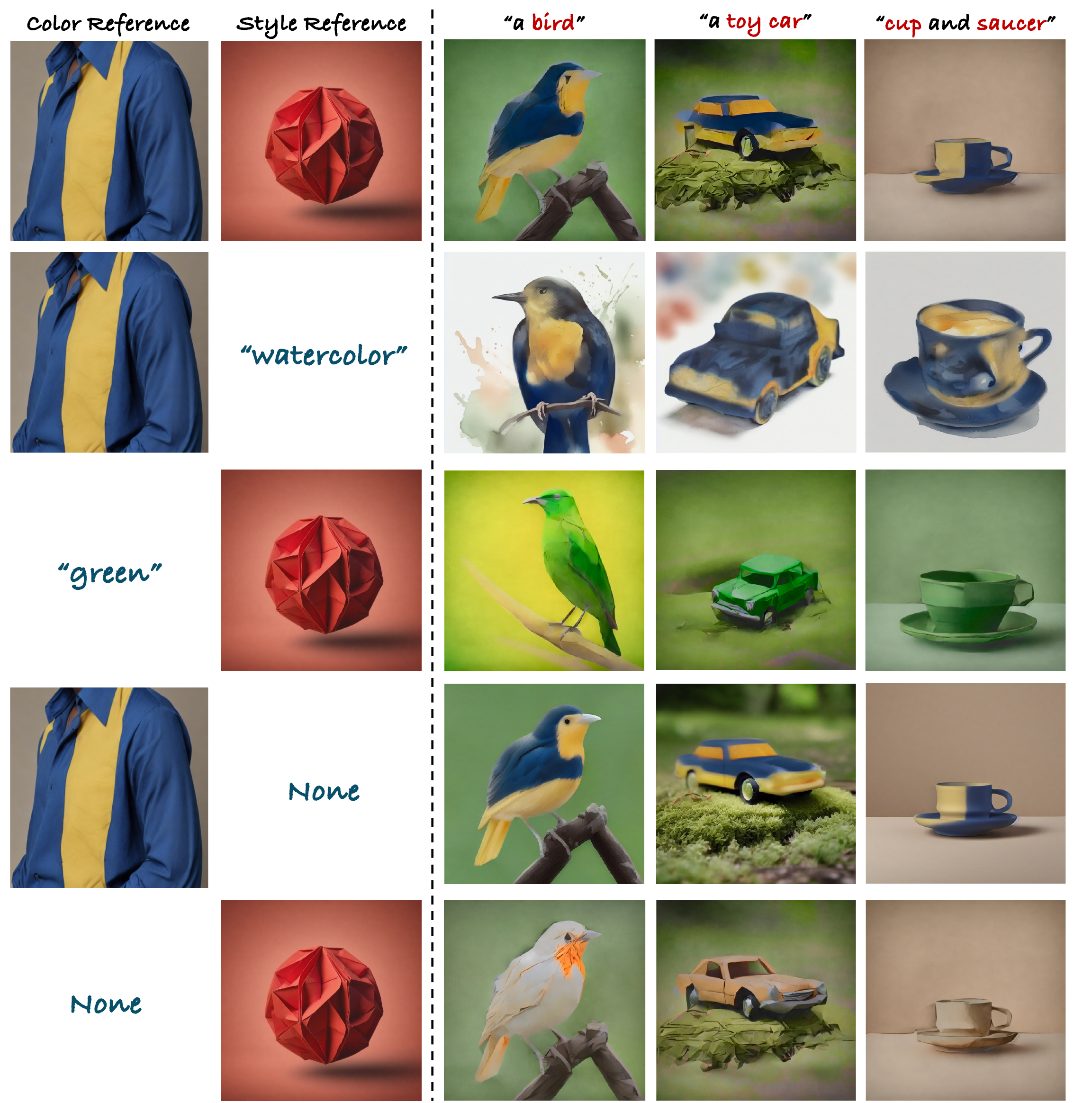}
 \vspace{-5pt}
 \caption{We propose the first training-free approach to allow disentangled conditioning of text-to-image diffusion models on color and style attributes from reference images.}
 \label{fig:teaser_qual}
\end{center}
}]

\ificcvfinal\thispagestyle{empty}\fi

\begin{abstract}
We consider the problem of independently, in a disentangled fashion, controlling the outputs of text-to-image diffusion models with color and style attributes of a user-supplied reference image. We present the first training-free, test-time-only method to disentangle and condition text-to-image models on color and style attributes from reference image. To realize this, we propose two key innovations. Our first contribution is to transform the latent codes at inference time using feature transformations that make the covariance matrix of current generation follow that of the reference image, helping meaningfully transfer color. Next, we observe that there exists a natural disentanglement between color and style in the LAB image space, which we exploit to transform the self-attention feature maps of the image being generated with respect to those of the reference computed from its L channel. Both these operations happen purely at test time and can be done independently or merged. This results in a flexible method where color and style information can come from the same reference image or two different sources, and a new generation can seamlessly fuse them in either scenario (see Figure~\ref{fig:teaser_qual}).

\end{abstract}

\section{Introduction}
\label{sec:intro}

We consider the problem of conditioning the text-to-image class of diffusion models \cite{ho2020denoising, rombach2022high, sohl2015deep, song2020denoising} on color and style attributes extracted from a user-provided reference image. In particular, we want to independently control outputs of text-to-image models with either or both of these attributes, necessitating disentangled color and style conditioning. Furthermore, we seek to do so in a completely training-free and test-time-only fashion. This is practically an important problem since (a) such disentangled control means color and style information can now come from two different sources, and a new generation conditioned on them can be fused to produce an image with color from the first source and style from the second source, and (b) a test-time and training-free solution means one does not have to keep training models each time reference images change.

While there has been much work in customizing text-to-image generation \cite{gal2022image, ruiz2023dreambooth, kumari2023multi, voynov2023p+, zhang2023prospect, agarwal2023image} with reference images, most of these techniques lack explicit control over which attributes from the reference are to be reflected in the synthesized images. Further, while there have been some attempts at training-free customization approaches, they all focus on a specific aspect, e.g., appearance transfer \cite{alaluf2023cross} or style transfer \cite{hertz2023style} or ensuring subject consistency \cite{tewel2024training}. None of these training-free methods are able to achieve disentangled attribute transfer that we seek to achieve in our work. Next, training-based methods such as MATTE \cite{agarwal2023image} proposed a way to allow attribute-conditioned image synthesis but it needed (a) optimizing textual tokens that may take hours depending on compute and reference image, and (b) a separate custom loss function to achieve disentanglement between color and style. Some other training-based methods such as ProSpect \cite{zhang2023prospect}, while doing multi-attribute conditioning, are also not able to disentangle color and style despite training tokens.  Consequently, we ask, and answer affirmatively, two key questions- (a) can we achieve test-time-only conditioning of text-to-image models with color and style attributes from reference images? and (b) can we do (a) with disentangled control of color and style? 

\begin{figure*}[tb]
  \centering
  \includegraphics[width=0.9\textwidth]{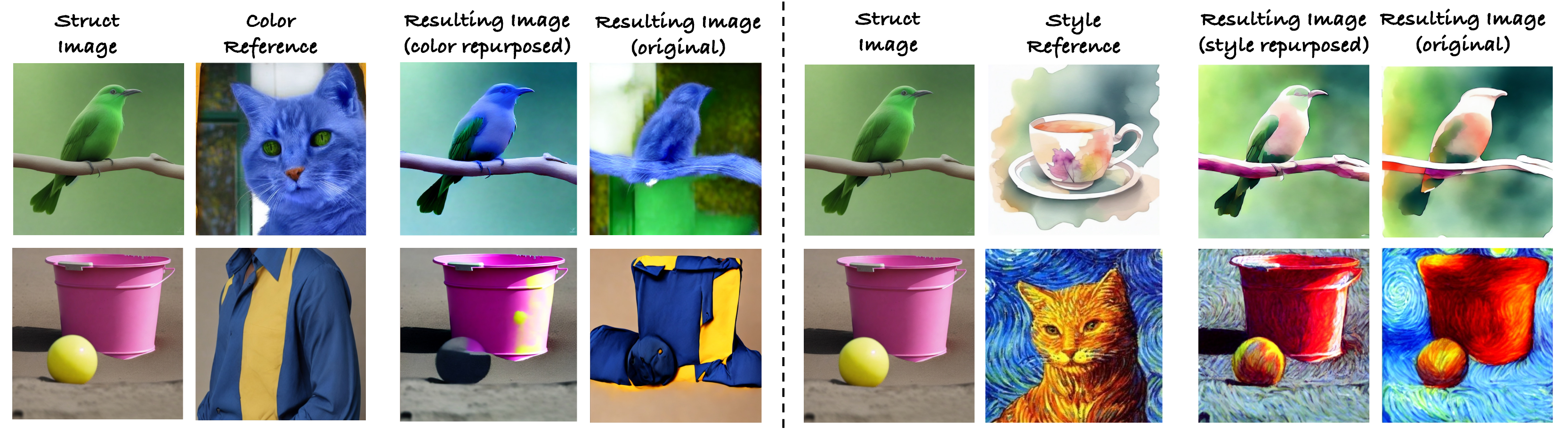}
  \caption{Attribute entanglement in prior appearance transfer works \cite{alaluf2023cross}.}
  \label{fig:cross_image_limn}
\end{figure*}

We begin with a brief discussion on why recent training-free methods such as \cite{alaluf2023cross} do not achieve disentangled attribute transfer. This method proposed to capture the customized concept by transfering keys and values computed from the reference image. Given observations from prior work \cite{zhang2023prospect} the color attribute is captured during the initial denoising stages and style in the later steps, a natural way to repurpose \cite{alaluf2023cross} for our task is to restrict these key-value operations to specific timesteps depending on the attribute we seek to transfer. We show some results with this approach in Figure~\ref{fig:cross_image_limn}. As can be noted from these results, the attribute transfers are far from desirable. This is because color transfer using the key-value operations of \cite{alaluf2023cross} is limited by the quality of semantic correspondences between the reference image and the current generation. On the other hand, transferring style by simply limiting key-value copy to the last denoising timesteps is insufficient since by then features would have sufficiently entangled color and style information. Consequently, it is critical to disentangle these attributes in the feature space in a principled fashion to be able to achieve multi-attribute transfer at test time.

\begin{figure}[h]
  \centering
  \includegraphics[width=0.3\textwidth]{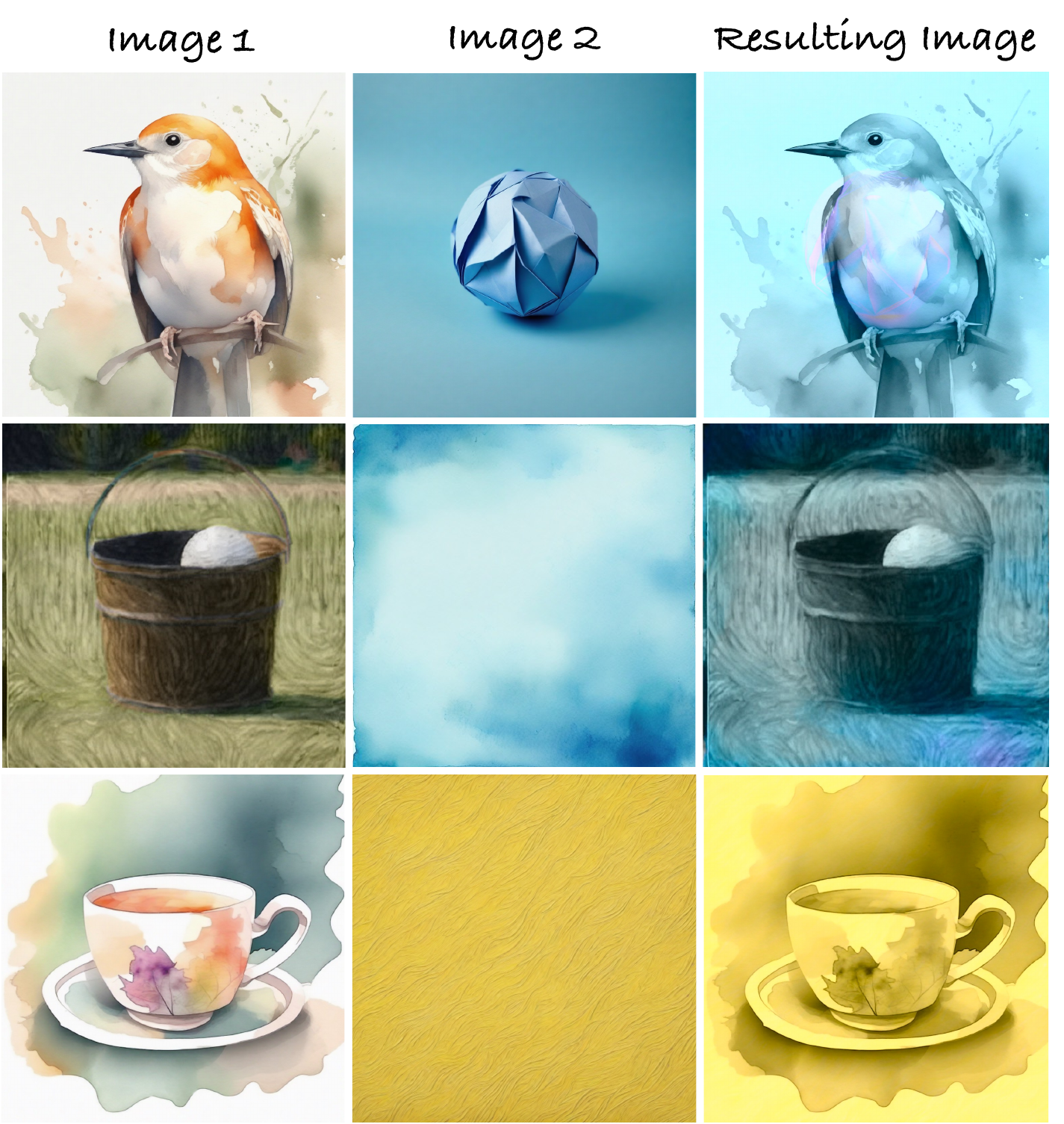}
  \caption{Attribute disentanglement in LAB space.}
  \label{fig:labRefs}
\end{figure}

To address the aforementioned issues, we propose the first training-free method to enable disentangled control over color and style attributes from a reference image when generating new images. To extract and transfer the color attribute from a reference image, we propose to modify the latent codes during denoising using a novel correspondence-aware recoloring transformation. Our key intuition is there naturally exist color clusters in the reference image, and ensuring regions with the dominant color populations from reference correspond to regions in the image being generated will lead to meaningful color transfer. Given a color clustering of the reference image, we realize this by picking a certain denoising timestep, decoding the latent code, clustering the image, establishing a correspondences between the two sets of clusters, and performing a correspondence-aware whitening and recoloring feature transformation on the latent codes. At the end of the denoising process, the final latent code when decoded will give a new image with colors from the reference image. For instance, see first row/first column in Figure~\ref{fig:teaser_qual} where the bird follows the colors of the shirt. Next, to transfer style and disentangle it from the color attribute, we propose two innovations. First, we observe that there exists an inherent disentanglement between these two attributes in the LAB color space where the L channel contains content and style information and the A/B channels the color information \cite{10.1145/3394171.3413853}. See Fig~\ref{fig:labRefs} where we also show this qualitatively. In each row, we take the L channel from the image shown in the first column (after converting the RGB image shown into LAB), the A/B channels from the second column, and merge them, giving the result in the third column. In the first row, we see the style from the first column and the colors (blue) from the second column get captured in the resulting bird image. Similarly, in the third row, despite the second column having certain textured patterns, only the color (yellow) gets transferred to the resulting image. Next, noting that style mostly gets captured during the later denoising steps, we propose a time-step-constrained feature manipulation strategy. We do this by first generating an image with the baseline model and the desired prompt, and store the A/B channels. We then transfer style using the L channel from the reference image by aligning the self-attention key-value feature maps of the reference with those of the current generation, copy the A/B channels from the above operation and obtain the final result. See some results with our method in Figure~\ref{fig:teaser_qual} where in each case, our method is able to respect both color and style references in the final outputs. For instance, the first row has a ``a bird" image following the blue/yellow colors from the color reference and origami style, the second row has the bird in blue/yellow colors and watercolor style, and so on.

To summarize, our key contributions in this work are:
\begin{itemize}
    \item We present the first training-free method to disentangle and control text-to-image diffusion models on color and style attributes from a reference image.
    \item We propose a new time-step-constrained latent code recoloring transformation that aligns the covariance matrices of a text-to-image model output with that a reference image, helping transfer reference colors to outputs of text-to-image models.
    \item We notice that the L channel in the LAB space has an inherent separation between style and color and propose a new time-step-constrained self-attention key and value feature manipulation algorithm to transfer style from a reference image.

\end{itemize}

\section{Related Work}
With the wide adaptation of diffusion models for text-to-image synthesis \cite{nichol2021glide, rombach2022high, ramesh2022hierarchical, saharia2022photorealistic}, much recent effort has been expended in controlling the outputs of these models. These efforts largely focus on learning adapters \cite{zhang2023adding,mou2023t2i,ye2023ip,sohn2023styledrop} given baseline text-to-image models, finetuning parameters of the base model\cite{ruiz2023dreambooth,kumari2023multi,tewel2023key,xiao2023fastcomposer}, learning new tokens in the vocabulary of these models \cite{gal2022image, agarwal2023image, zhang2023prospect, zhang2023inversion}, introducing dedicated personalization encoders \cite{gal2023encoder,li2024blip,shi2024instantbooth,wei2023elite}, utilising LoRA \cite{hu2021lora} to encapsulate target information \cite{li2024diffstyler, yang2024lora,gu2024mix,shah2023ziplora,wang2023autostory,zhong2024multi}, and inpainting-based approaches \cite{yang2023paint,chen2024anydoor}. With the exception of MATTE \cite{agarwal2023image}, none of these methods are able to achieve disentangled transfer between color and style of a reference image. However, MATTE \cite{agarwal2023image} needs custom loss functions and hours of training per reference image, which is practically infeasible. 

On the other hand, there is a new line of recent work that involves training-free approaches to customization \cite{yu2023freedom,alaluf2023cross,hertz2023style,tewel2024training,chung2024style,xu2024freetuner,wang2024instantstyle,wang2024instantstyle2,chen2024anydoor}. However, these methods focus on one specific aspect (e.g., appearance transfer, style transfer, or subject consistency) and are not able to provide independent disentangled control over color and style attributes.  Similarly, the non-diffusion based conventional style transfer methods AesPA-Net\cite{hong2023aespa}, StyTR2\cite{deng2022stytr2}, and ArtFlow\cite{an2021artflow} target disentanglement of content and style, while style and color are not treated independently/separately in these works. We address these gaps in both training-based and training-free methods by proposing the first training-free method to provide disentangled control for text-to-image models over color and style attributes from reference images. Moreover, our proposed approach is not specific to any dataset, and can seamlessly adapt to various styles/content due to the base model's capability.

\section{Approach}
\label{sec:approach}
We start with a brief review of latent diffusion models (LDMs). LDMs comprise an encoder-decoder pair and a separately trained denoising diffusion probabilistic model (DDPM). Leveraging an encoder $\mathbf{E}$, LDMs translate an image $\mathbf{I}$ into a latent code $\mathbf{z}$, perform iterative denoising, and subsequently convert the predicted latent codes back to the pixel space via the decoder $\mathbf{D}$. The training objective of the DDPM $\epsilon_{\theta}$ is the following: $\mathbb{E}_{\mathbf{z}\sim \mathbf{E}(\mathbf{I}),p,\epsilon\sim\mathcal{N}(0,1),t}[\|\epsilon - \epsilon_{\mathbf{\theta}}^{(t)}(\mathbf{z}_{t}, \mathbf{L}(p))\|] $
where p denotes any external conditioning factor e.g., a text prompt, which is typically encoded using text encoder $\mathbf{L}$ (e.g., CLIP \cite{radford2021learning}, T5 \cite{raffel2020exploring}). At any timestep $t$ of the denoising process, given the current latent code $z_{t}$, the goal is to produce $z_{t-1}$.  The first step here is to predict the noise $\epsilon_\theta^{(t)}(\mathbf{z}_{t}, \mathbf{L}(p))$. Given $z_{t}$ and $\epsilon_\theta^{(t)}(\mathbf{z}_{t}, \mathbf{L}(p))$, deterministic DDIM \cite{song2020denoising} sampling gives $z_{t-1}$ as

\begin{equation}
    \label{eqn: z_comp}
    z_{t-1} =  \sqrt {\alpha_{t-1}} z_0 + \hat{x}_t
\end{equation}
where $z_0$ (denoised prediction) is predicted as $z_0 = \frac {z_{t}-{\sqrt {1-{\bar {\alpha }}_{t}}}\epsilon_\theta^{(t)}}{\sqrt {{\bar {\alpha }}_{t}}}$, and $\hat{x}_t$ (direction pointing to $x_t$) is computed as $\hat{x}_t = \sqrt{1-\alpha_{t-1}-\sigma_t^2}\epsilon_\theta^{(t)}$.

\subsection{Disentangled Color and Style Conditioning}

\begin{figure*}[tb]
  \centering
  \includegraphics[width=0.8\textwidth]{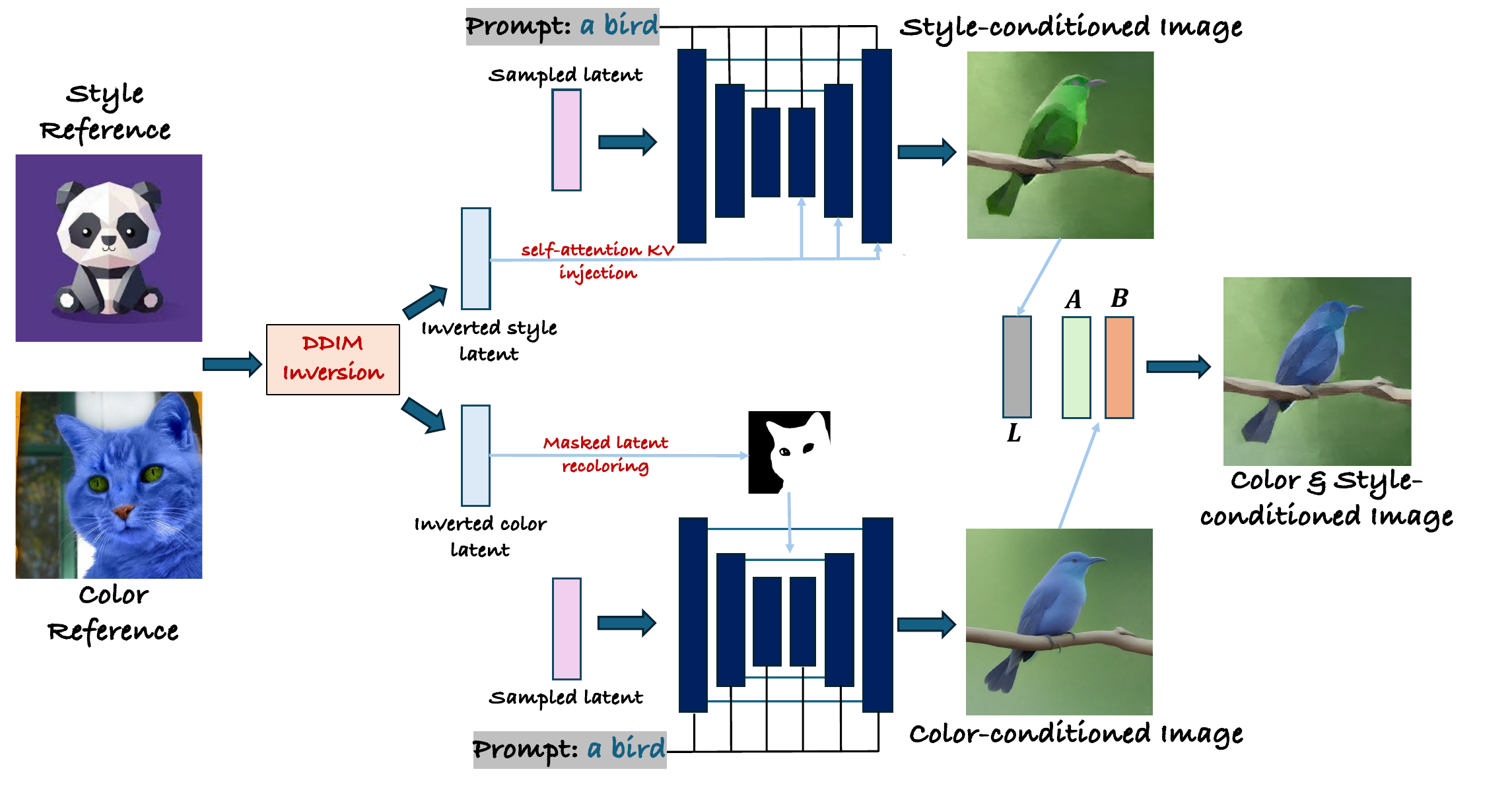}
  \caption{A visual illustration of the proposed method. (a) Obtain inverted latents for the style and color reference images, (b) Sample a new latent given a prompt (e.g., \texttt{a bird} above) and begin the denoising process for generating new images, (c) Perform \textit{self attention KV injection} for style transfer, and/or \textit{masked latent recoloring} for color transfer during this ongoing denoising process, (d) Utilise the intermediates obtained from style/color reference image reconstruction via style/color inverted latents respectively in step (c)}
  \label{fig:overall_pipeline}
\end{figure*}

As discussed in Section~\ref{sec:intro}, we seek to provide text-to-image models with disentangled control over \texttt{color} and \texttt{style} attributes extracted from user-supplied reference images. To do this, we  propose a new training-free algorithm that facilitates any of color-only, style-only, or both color-style transfer from reference images. We achieve this with a two-branch architecture (see Figure \ref{fig:overall_pipeline}), one each for color and style. As we discuss later, outputs from these branches can be used independently (for single attribute transfer) or can be merged seamlessly. This merging can happen with color and style from the same source (one reference image) or color from one image and style from another image (see Fig~\ref{fig:teaser_qual} again where we show both single-source and two-source results).

Our closest training-free baselines \cite{alaluf2023cross, tewel2024training} inject key and value feature maps from self-attention blocks of the U-Net from the reference image. However, this only helps transfer the overall appearance/identity and cannot control color and style. To fix this gap, we propose two ideas. First, given color information from a reference image, we propose to apply recoloring transformations on the intermediate latent codes which when decoded can give an image following reference colors. Next, we notice that (a) style information gets captured only during the later parts of the denoising process and (b) style is captured in the L channel when an image is converted from the RGB to the LAB space (see Section~\ref{sec:intro} and Fig~\ref{fig:labRefs} again). We exploit these two observations to propose a timestep-constrained key and value feature manipulation strategy in the L space where features during early denoising steps are retained as is from the baseline generation and those of later steps are carried over from the reference image. We next explain the details of these ideas.

\begin{figure*}[tb]
  \centering
  \includegraphics[width=0.9\textwidth]{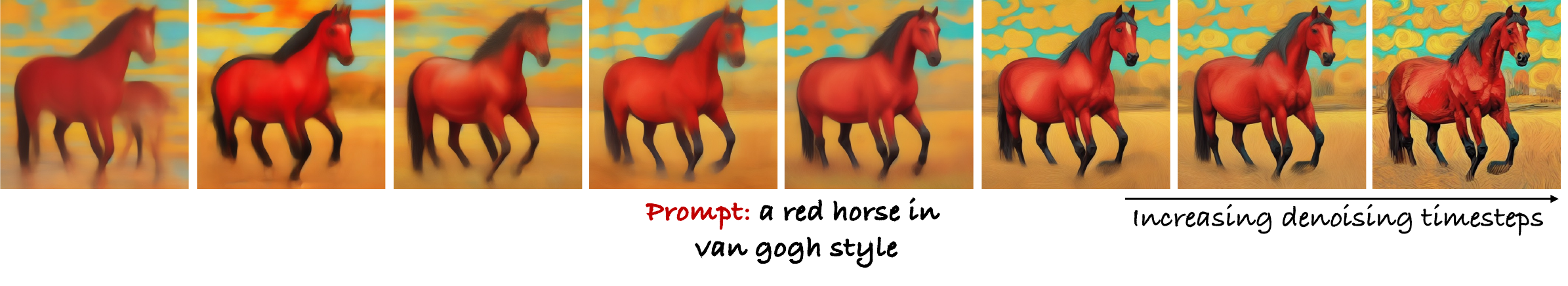}
  \vspace{-6pt}
  \caption{Intermediate decoded latents demonstrate the progression of color and fine-grained style information across denoising timesteps.}
  \label{fig:example}
\end{figure*}

\textbf{Color conditioning.}
Our proposed method is visually summarized in the \texttt{color} branch in Figure \ref{fig:overall_pipeline}. We first do a DDIM inversion step on the reference image to obtain the corresponding latent $z_t^{\text{ref}}$. Once the denoising process begins given a user-specified text prompt, given a latent $z_t$ at timestep $t$, the DDIM sampling will compute the noise prediction $\epsilon_\theta^{(t)}$, followed by computing the $z_0$ for both the reference image and the new generation. We then decode the latent code $z_0$ using the decoder $\mathcal{D(.)}$ (see Figure \ref{fig:example} for an example where we visualize these for several intermediate denoising timesteps). Given a timestep t, we first perform a K-Means clustering operation on both the decoded image $I_0^{(t)_{gen}}$ and the reference image to obtain sets of $K$ color clusters $\mathcal{C}_{\text{gen}}$ and $\mathcal{C}_{\text{ref}}$ respectively. Note that one can mask the decoded latent with cross-attention maps to restrict the object of interest in both the new generation as well as the reference image. Given the cluster sets, we next establish correspondences between them based on their proportion, giving a set of masks $\mathcal{M}_{ref}$ and $\mathcal{M}_{gen}$ for each set. The idea here is that a color cluster with the largest membership in the reference image indicates the dominant color that we seek to transfer to the current generation.  
Given the masks above, we achieve this by applying a mask-aware recoloring transformation (RT) on the latent code $z_0^{(t)_{gen}}$ as:

\begin{align}
\label{eqn: wct}
    z_0^{(t)_{gen}} = & \sum_{\substack{ m_{gen}^i \in \mathcal{M}_{gen} \\ m_{ref}^i \in \mathcal{M}_{ref} \\  1 \leq i \leq k}} \Bigg\{ \left( 1-m_{gen}^i \right) z_0^{(t)_{gen}} \nonumber \\
    & + m_{gen}^i \left[\text{RT}\left(m_{gen}^i z_0^{(t)_{gen}}, m_{ref}^i z_0^{(t)_{ref}} \right) \right] \Bigg\}
\end{align}

Here, we iterate over all the $K$ clusters and apply the recoloring transform separately to regions determined by masks corresponding to each cluster. In each iteration $i$, we use the corresponding mask $m_{gen}^i$ to constrain the region where color transfer happens, and similarly $m_{ref}^i$ determines the reference pixels from where the colors are picked. This way, we ensure that in any iteration $i$, pixels outside the region of interest (determined by the $m_{gen}^i$) remain untouched. The recoloring transformation \cite{hossain2014whitening} itself is a two-step process. We first whiten the latent codes to get its covariance matrix to be identity. We then apply a transformation to match the covariance matrix of the latent codes to match that of the reference image ($z_0^{(t)_{ref}}$). To ensure this operation strictly transfers color only and not style, we use observations from prior work \cite{agarwal2023image,zhang2023prospect} that note color is captured during the early parts of the denoising process. Consequently, we restrict Eq~\ref{eqn: wct} to only a subset of the initial denoising timesteps $t_{start}^c>t>t_{end}^c$.

\begin{figure*}
  \centering
  \includegraphics[width=0.85\textwidth]{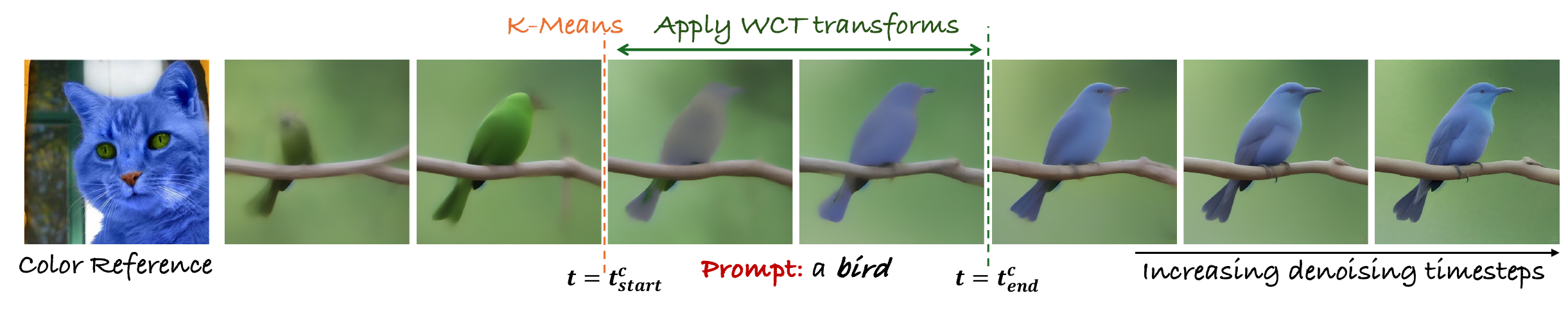}
  \vspace{-6pt}
  \caption{Progression of recolorised decoded latents across denoising timesteps.}
  \label{fig:color_transfer_progression}
\end{figure*}

The updated $z_0^{(t)_{gen}}$ obtained in Equation \ref{eqn: wct} is then used along with the predicted noise $\epsilon_\theta^t$ to compute $z_{t-1}^{gen}$ (using Equation \ref{eqn: z_comp}) which goes as input to the next denoising step, eventually generating an image which follows the color from the reference image. 
We show an example demonstrating the progression of decoded latents $I_0^{(t)_{gen}}$ across denoising timesteps in Figure \ref{fig:color_transfer_progression}. Here given the prompt \texttt{a bird}, one can observe that the model very initially starts forming some colors (\texttt{green} here). We transform the intermediate latents to manipulate these colors using the steps described above and obtain a \texttt{blue} bird following the blue cat in the color reference image shown in the figure.

\textbf{Style conditioning.} Given that high-frequency details like \texttt{style} and \texttt{texture} shows up in the later denoising timesteps \cite{agarwal2023image, zhang2023prospect}, we begin by injecting key and value feature maps from the reference image to the current generation for only the last few ($t > t^s_{start}$) denoising timesteps. However, an issue with this approach is the feature maps in the later timesteps will have color information as well (since color would have been captured in the beginning), leading to entanglement of color and style. See the ``Resulting image (style repurposed)" column in Fig~\ref{fig:cross_image_limn} for results with this approach- clearly both style and colors are getting transferred in this case, e.g., bird has both the watercolor style and the light pink colors from the ``reference" image. To be able to disentangle style from color and allow independent control of the text-to-image diffusion model, our key insight is that there exists an inherent separation between style and color in the LAB space. The L channel captures the content and style, whereas the AB channels have color information (recall our discussion of Figure~\ref{fig:labRefs} in Section~\ref{sec:intro}). Since diffusion models are generally trained to operate in the RGB space, we take the grayscale version of the reference as an approximation to the L channel for all the next steps below (see Fig~\ref{fig:overall_pipeline} again for a visual summary). We begin by DDIM inverting the reference to get the latent $z_t^{ref}$. Given any user-specified text prompt (e.g., \texttt{a bird}), for each denoising timestep $t < t^s_{start}$, we denoise the input latent codes as in a baseline text-to-image model but once we hit $t > t^s_{start}$, we start injecting the self-attention key $K$ and value $V$ feature maps from the reference reconstruction after converting it to grayscale and DDIM inverting it as noted above. Formally, this modified self-attention feature map computation at any denoising timestep $t$ and layer $l$ of the U-Net can be expressed as:

\begin{align}
    \hat{f}_t^l = & \, \mathbbm{1}_{0 \leq t < t^s_{\text{start}}} \, \text{softmax} \left( \dfrac{Q_{gen}^l K_{gen}^l{}^T}{\sqrt{d_k}} V_{gen}^l \right) \nonumber \\
    & + \mathbbm{1}_{t > t^s_{\text{start}}} \, \text{softmax} \left( \dfrac{Q_{gen}^l K_{ref}^l{}^T}{\sqrt{d_k}} V_{ref}^l \right)
\end{align}

where $\mathbbm{1}$ is an indicator, and $Q_{gen}^l\slash K_{gen}^l\slash V_{gen}^l$ and $Q_{ref}^l\slash K_{ref}^l\slash V_{ref}^l$ denote $l^{th}$ U-Net layer self-attention queries, keys, and values for the generation and reference respectively. We then take the final latent code, decode it to get an image, convert it to the LAB space, retain the L channel and get the AB channels from the corresponding color branch of Figure~\ref{fig:overall_pipeline}.

Note that $t_{end}^c$ ($=T/5$) is strictly less than $t>t_{start}^s$ ($=4T/5$) throughout (i.e. the timestep intervals for which color and style conditioning is applied have no overlap), where $T$ denotes the total number denoising steps.

\section{Results}

\begin{figure}[tb]
  \centering
  \includegraphics[width=0.45\textwidth]{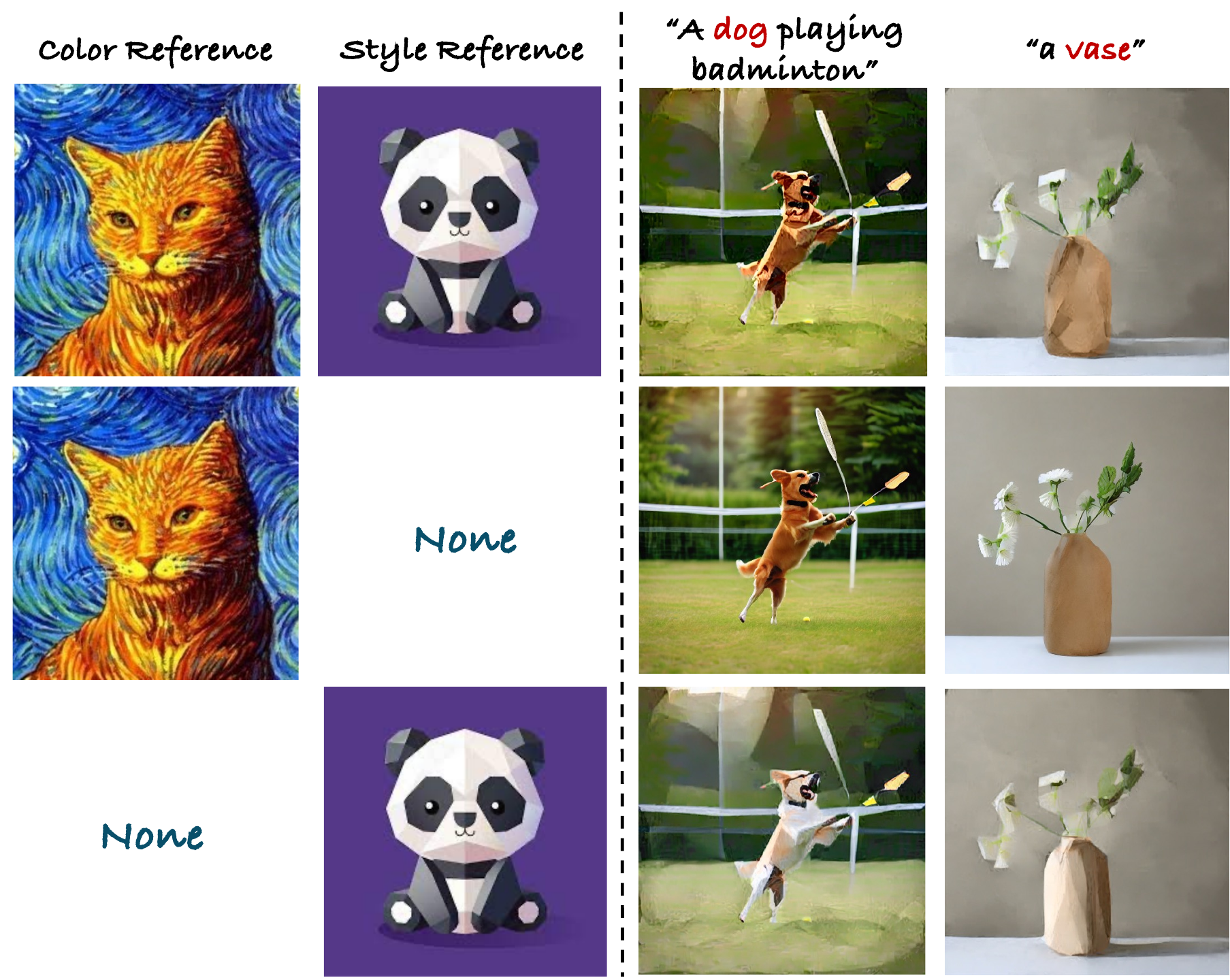}
  \vspace{-6pt}
  \caption{Disentangled color-style transfer from reference images.} 
  \label{fig:ours_results}
\end{figure}

\textbf{Qualitative Evaluation.} In addition to the results in Figure \ref{fig:teaser_qual}, we show more results with our method in Figure \ref{fig:ours_results} (color and style reference in the first two columns and our results in columns three-four) to demonstrate disentangled transfer of color and style attributes from reference images. We show different combinations of control over color and style attributes. In the first row, we are able to generate images following the content from the prompt (\texttt{dog}, \texttt{vase}) while following the style and color from the provided reference images.  
In the last two rows, our method generates images following style or color from the reference while imposing no control over the other attribute (observe the straight and sharp edges in dog's outline in the last row).

\begin{figure*}[!ht]
  \centering
  \includegraphics[width=0.8\textwidth]{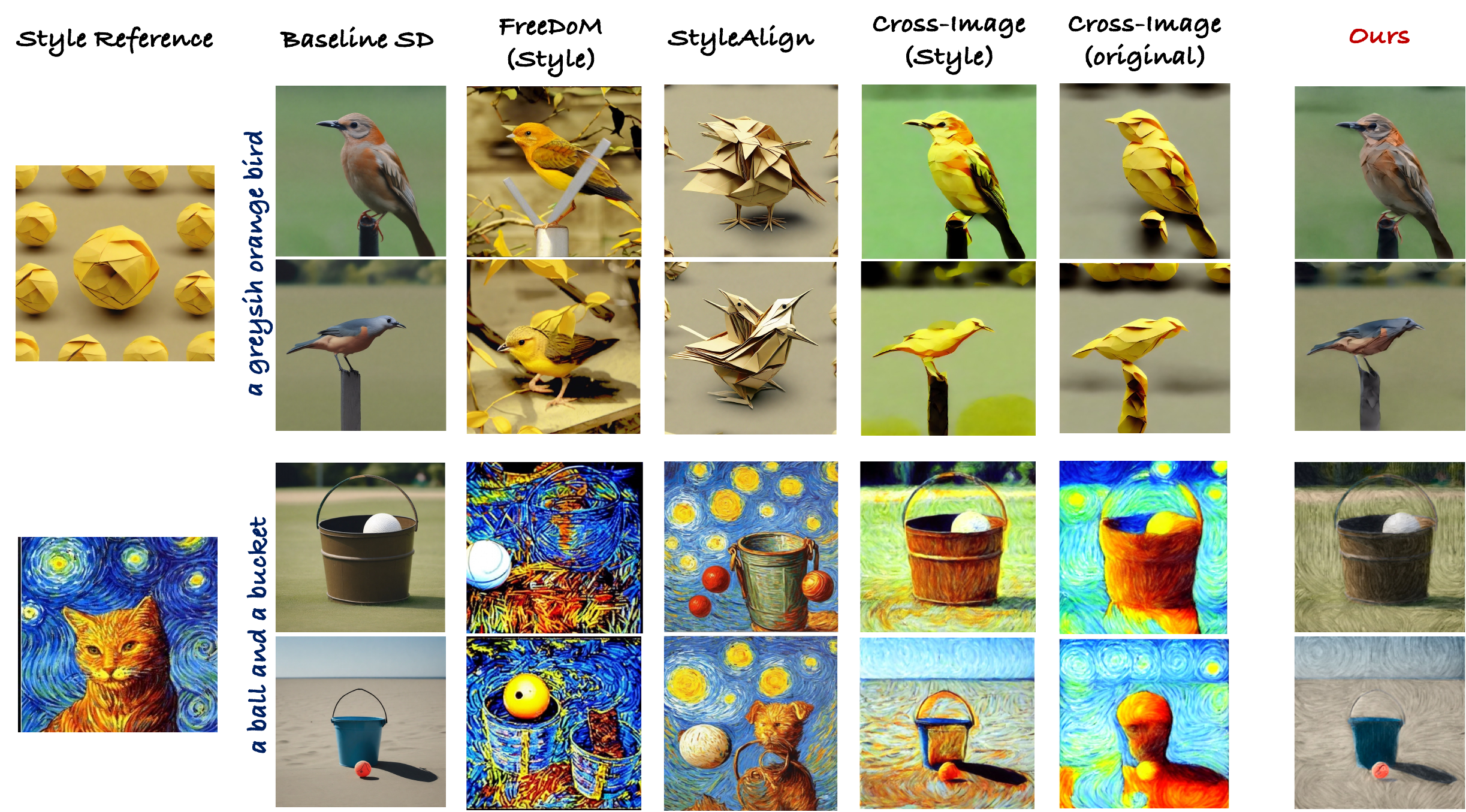}
\vspace{-6pt}
  \caption{Comparing with state-of-the-art methods for disentangled style transfer. Please zoom in for viewing details.} 
  \label{fig:style_results}
\end{figure*}

\begin{figure}[!ht]
  \centering
  \includegraphics[width=0.45\textwidth]{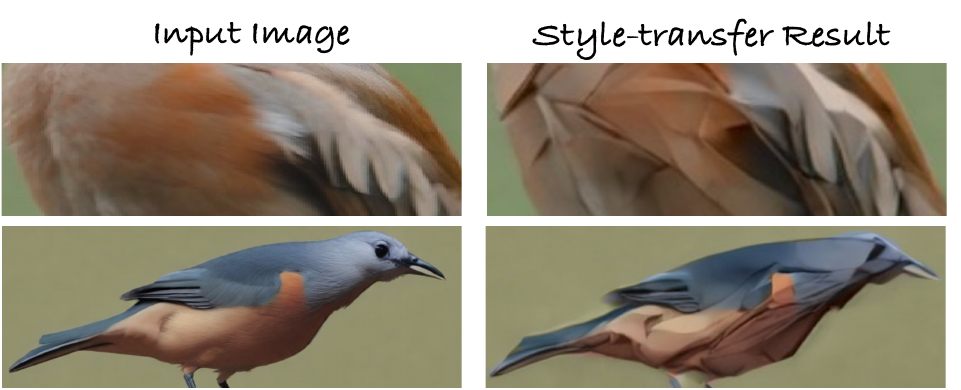}
\vspace{-6pt}
  \caption{Style transfer (Origami).} 
  \label{fig:STzoom}
\end{figure}

\begin{figure}[!ht]
  \centering
  \includegraphics[width=0.5\textwidth]{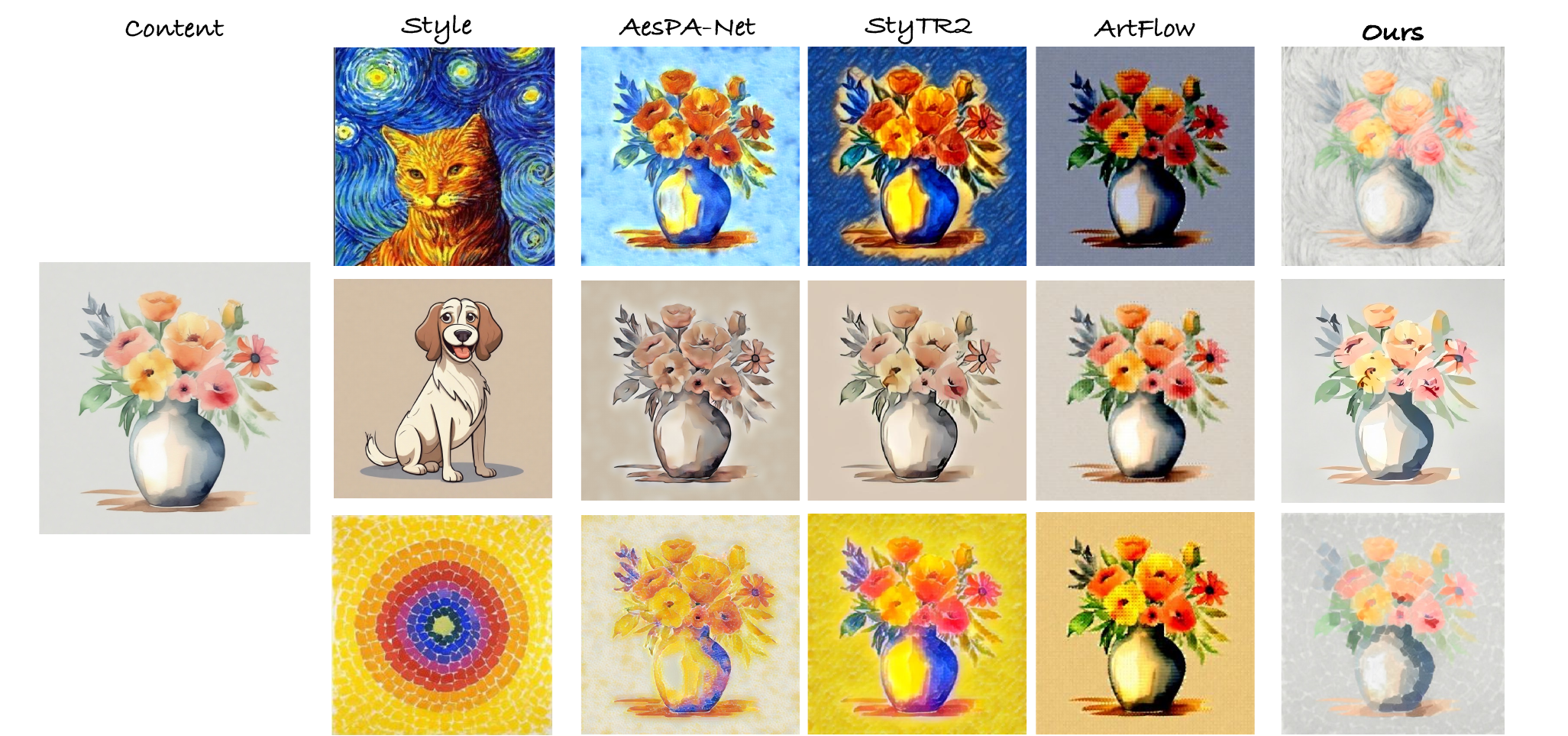}
  \vspace{-16pt}
  \caption{Comparison with conventional style-transfer baselines.} 
  \label{fig:NonDiffusionST}
\end{figure}

\begin{figure}[!ht]
  \centering
  \includegraphics[width=0.45\textwidth]{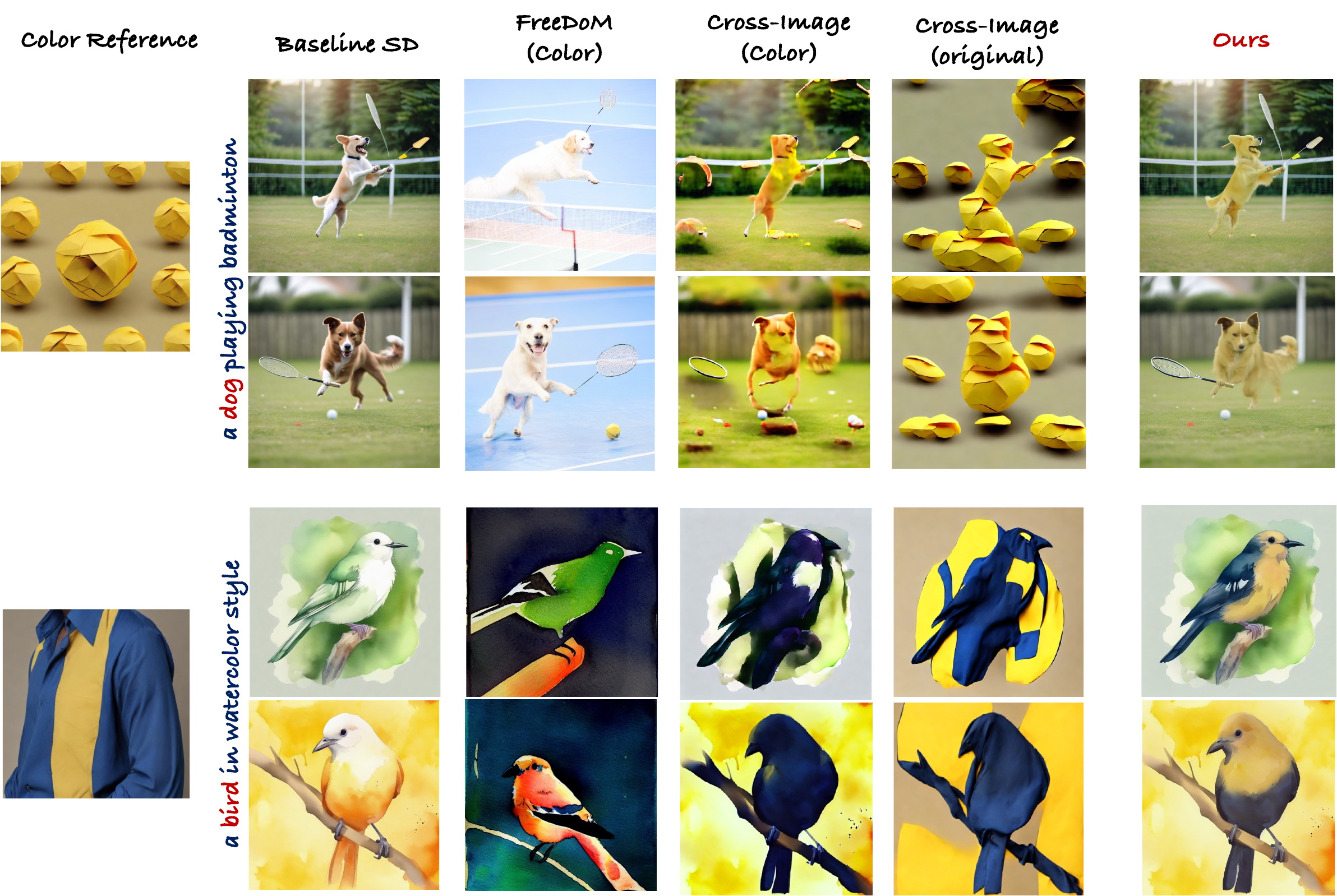}
  \vspace{-6pt}
  \caption{Comparing with state-of-the-art methods for disentangled color transfer.} 
  \label{fig:color_results}
\end{figure}

\begin{figure}[!ht]
  \centering
  \includegraphics[width=0.4\textwidth]{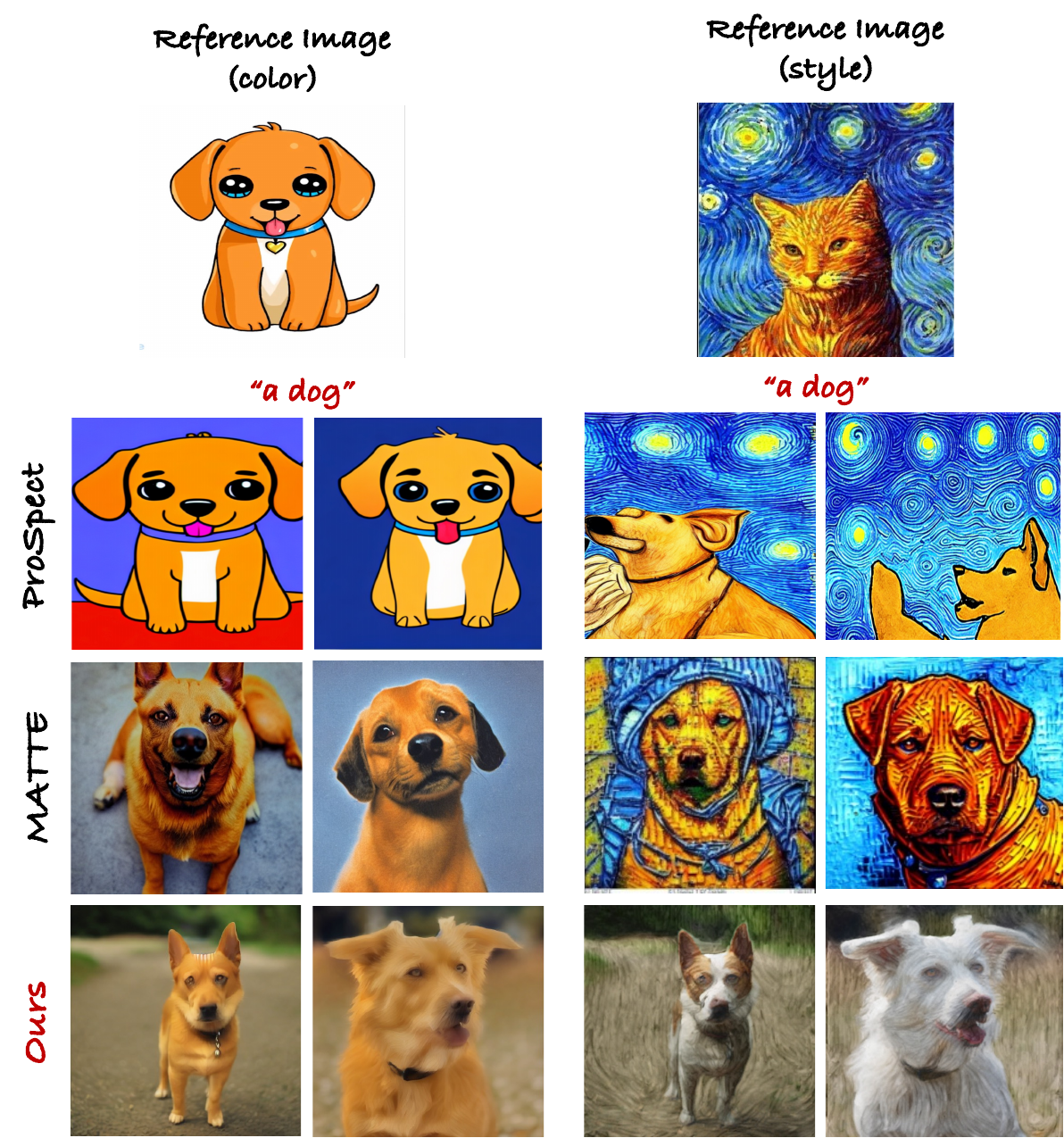}
  \vspace{-6pt}
  \caption{Comparing with recent state-of-the-art methods for multi-attribute constrained generation. Please zoom in for details.} 
  \label{fig:baseline_comp_training}
\end{figure}

We next compare our method with various baselines including training-free style transfer (Cross-Image \cite{alaluf2023cross}, StyleAlign \cite{hertz2023style}, FreeDoM \cite{yu2023freedom}) in Figure \ref{fig:style_results}, conventional diffusion-free style transfer baselines (AesPA-Net\cite{hong2023aespa}, StyTR2\cite{deng2022stytr2}, and ArtFlow\cite{an2021artflow} in Figure~\ref{fig:NonDiffusionST}), training-free color transfer (Cross-Image \cite{alaluf2023cross}, FreeDoM \cite{yu2023freedom}) in Figure \ref{fig:color_results}, and training-based techniques such as MATTE \cite{agarwal2023image} and ProSpect \cite{zhang2023prospect} in Figure \ref{fig:baseline_comp_training}.

We begin by discussing disentangled style transfer results in Figure \ref{fig:style_results}. One can note that our method is able to generate images (see last column) following the style from the reference image in a disentangled manner without affecting any other aspect of what the baseline generates (See Figure~\ref{fig:STzoom} for zoomed in section of images for the first/second rows of origami style transfer). On the other hand, as expected, the Cross-Image baseline transfers the full appearance of the reference image. One can also notice missing regions in the bucket image in fourth row/fifth column due to lack of semantic correspondences. Similarly, the repurposed style version of cross-image baseline, while doing better, can not fully disentangle style from color as the features already have color information by the time style transfer happens during denoising. In StyleAlign, in addition to the style and color being entangled, the algorithm also transfers the structural/layout aspects of the reference image, thereby limiting the kind of control with style we seek over the final outputs (see fourth row/third column, where it tries generating bucket images following the layout of the cat from the reference image). Finally, FreeDoM also entangles color and style because its loss function does not account for any explicit disentanglement of these attributes.

We also compare with conventional style transfer methods in Figure~\ref{fig:NonDiffusionST}. The baselines are able to preserve input content, but the transfer of style (e.g. van-gogh patterns in first row) is limited. Since these methods target disentanglement of content and style, style and color are not treated independently/separately, leading to undesirable results (e.g. see the first row/second-third columns where the result flower vase has blue/orange colors from the style reference whereas our method in the last column is able to fix this issue). Finally, while these methods are trained on specific datasets, our method is training free and not specific to any dataset, and can seamlessly adapt to various styles/content due to the base model's capability.

We next discuss disentangled color transfer results in Figure \ref{fig:color_results}. In all cases, our method is able to correctly transfer the color from the reference image whereas the cross-image (original) baseline transfers the full appearance from the reference image and can not control the color attribute independently (see third/fourth rows where the generated birds have a cloth-like appearance). Similarly, the repurposed color version of cross-image baseline, while disentangling color and style, is unable to produce good results due to lack of correspondences. 
With FreeDoM, in some cases there is overfitting to the colors during optimization while disregarding the overall aesthetics, whereas in several other cases, the generated images disregards reference colors (the birds in last row/second column) due to incorrect optimization. Our method is able to control and transfer color attribute independently without affecting any other aspects of what the pretrained model would have generated. 

Finally, in Figure \ref{fig:baseline_comp_training}, we compare to recent training-based methods. In ProSpect \cite{zhang2023prospect}, one can see color and style are completely entangled (e.g., first column where both color and style are transferred). On the other hand, despite our method being completely training free, it performs at par with MATTE \cite{agarwal2023image} which is a training-based approach (e.g., first column with our orange dogs). In the second column, whereas both MATTE and ProSpect entangle color and style, our method is able to generate van gogh-styled dogs without the bluish-orange colors.

\textbf{Quantitative Evaluation.} We next quantify improvements with our proposed method. We wish to evaluate how well these methods disentangle style and color (we follow the protocol from MATTE \cite{agarwal2023image}), while also following the details specified as part of the text prompt. We keep either of the attributes (out of style/color) fixed from a reference image (we use the same set as in \cite{gal2022image, voynov2023p+, zhang2023prospect, agarwal2023image}) and vary the other (we use the list of $7$ \texttt{style} types, $13$ \texttt{object} types and $11$ \texttt{color} types from previous works \cite{voynov2023p+, agarwal2023image}). In each case, we synthesize a set of $64$ images and compute the average CLIP image-text similarity. A higher score indicates better disentanglement since both attributes would then be separately captured well in the output. 

To further evaluate the quality of transfer of each attribute, we also compute similarity scores between the ground truth color/style (color obtained using ColorThief \cite{dhakar2015color}) and the generated images. As can be seen from Table 1, our method outperforms all training-free baselines and allows for independent control over style and color attributes. Further, when compared to the training-based MATTE \cite{agarwal2023image}, our method performs very competitively despite being training free. 
Finally, we conduct a user study with the generated images where we show survey respondents a textual prompt followed by color and style references, and ask them to select the images (among sets from four different methods) that best follow the provided constraints. From Table 2, our method's results are preferred by a majority of users, providing additional evidence for effectiveness of the proposed approach. 

\begin{table}
\centering
\scalebox{0.8}{
\begin{tabular}{@{}c|c|c|c@{}}
\toprule
Method                             & color & style & disentanglement \\ \midrule
ProSpect \cite{zhang2023prospect}  & 0.26 & 0.26& 0.19\\
MATTE \cite{agarwal2023image}      & 0.27 & 0.27& 0.26\\
FreeDoM \cite{yu2023freedom}       & 0.20 & 0.24& 0.19\\
StyleAlign \cite{hertz2023style}   & 0.26 & 0.25& 0.20\\
Cross-Image \cite{alaluf2023cross} & 0.25 & 0.26& 0.20\\ \midrule
\textbf{Ours}                      & 0.28 & 0.26& 0.26\\ \midrule

\end{tabular}
}
\vspace{-8pt}
\caption{Comparison with baselines.}
\end{table}

\begin{table}
\centering
\scalebox{0.8}{
\begin{tabular}{@{}c|c@{}}
\toprule
Method                             & Preference \\ \midrule
FreeDoM \cite{yu2023freedom} &  $4.9\%$\\
StyleAlign \cite{hertz2023style} &  $7.1\%$\\
Cross-Image \cite{alaluf2023cross} &  $18.7\%$\\ \midrule
\textbf{Ours} &  \textbf{69.3}\%\\ \midrule
\end{tabular}
}
\vspace{-8pt}
\caption{User study results.}
\end{table}

\section{Summary}

We considered the problem of disentangled color and style control of text-to-image models and noted none of the existing methods address this problem with a training-free approach. To this end, we proposed the first training-free, test-time-only solution with two key novelties: a timestep-constrained latent code recoloring transformation that aligned colors of generation outputs with reference colors and a timestep-constrained self-attention feature manipulation strategy in the L channel of the LAB space that aligned generation the style of generation outputs with that of the reference. This resulted in a flexible approach that can do color-only, style-only, or both color-style conditioning in a disentangled and indepedent fashion. Extensive qualitative and quantitative evaluations demonstrated the efficacy of our proposed method.

\begin{appendices}

\section{}
In Section~\ref{subsec:style_ref}, we validate our choice of using a grayscale version of the provided reference image compared to the colored reference image. In Section~\ref{subsec:additional_qual}, we show additional results for disentangled conditioned on color and style attributes from a user-provided reference image. In Section~\ref{subsec:color_results}, we show applicability of our proposed color conditioning in generating color variants of a given image. In Section~\ref{subsec:style_results}, we show additional results for transfer of style attribute from a reference image. Finally, we conclude with some discussion on
limitations of our method in Section~\ref{subsec:limitations}.

\subsection{Style conditioning with colored reference}
\label{subsec:style_ref}
\begin{figure*}[tb]
 \centering
 \includegraphics[width=\textwidth]{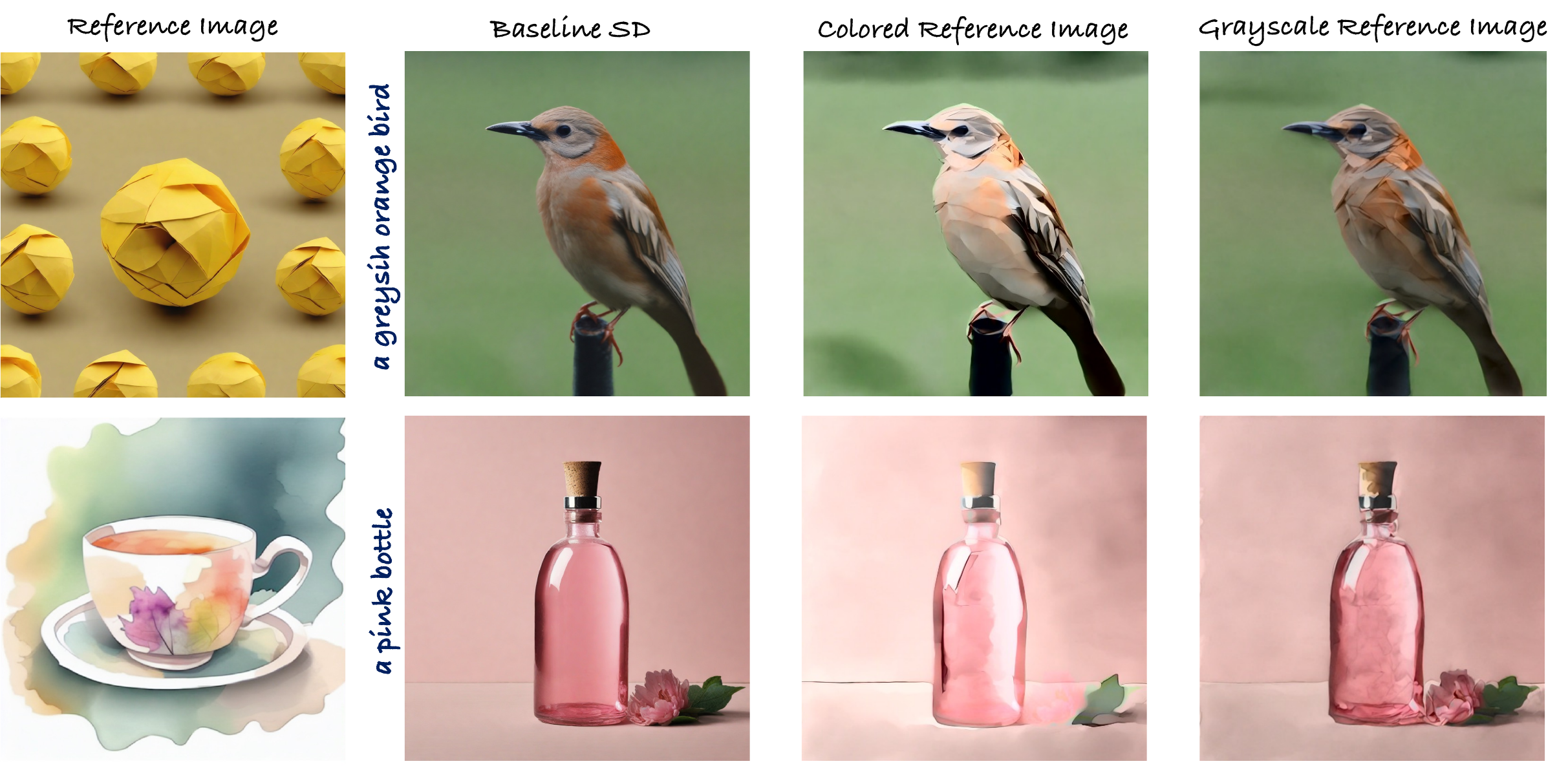}
 \caption{Comparing style transfer using original (colored) and grayscale reference image.} 
 \label{fig:style_colorgray_ref}
\end{figure*}

We experimented with taking the original (colorised) reference image instead of using its' grayscale counterpart. But we observed that the presence of color information in the reference image, leads to the intensity of colors being modified in the generated image, thereby leading to differences in lightness/darkness of some colors. We show some examples for the same in Figure \ref{fig:style_colorgray_ref}. This further validates our choice of using a grayscale reference while transferring style.

\subsection{Additional Qualitative Results}
\label{subsec:additional_qual}
\begin{figure*}[tb]
  \centering
  \includegraphics[width=\textwidth]{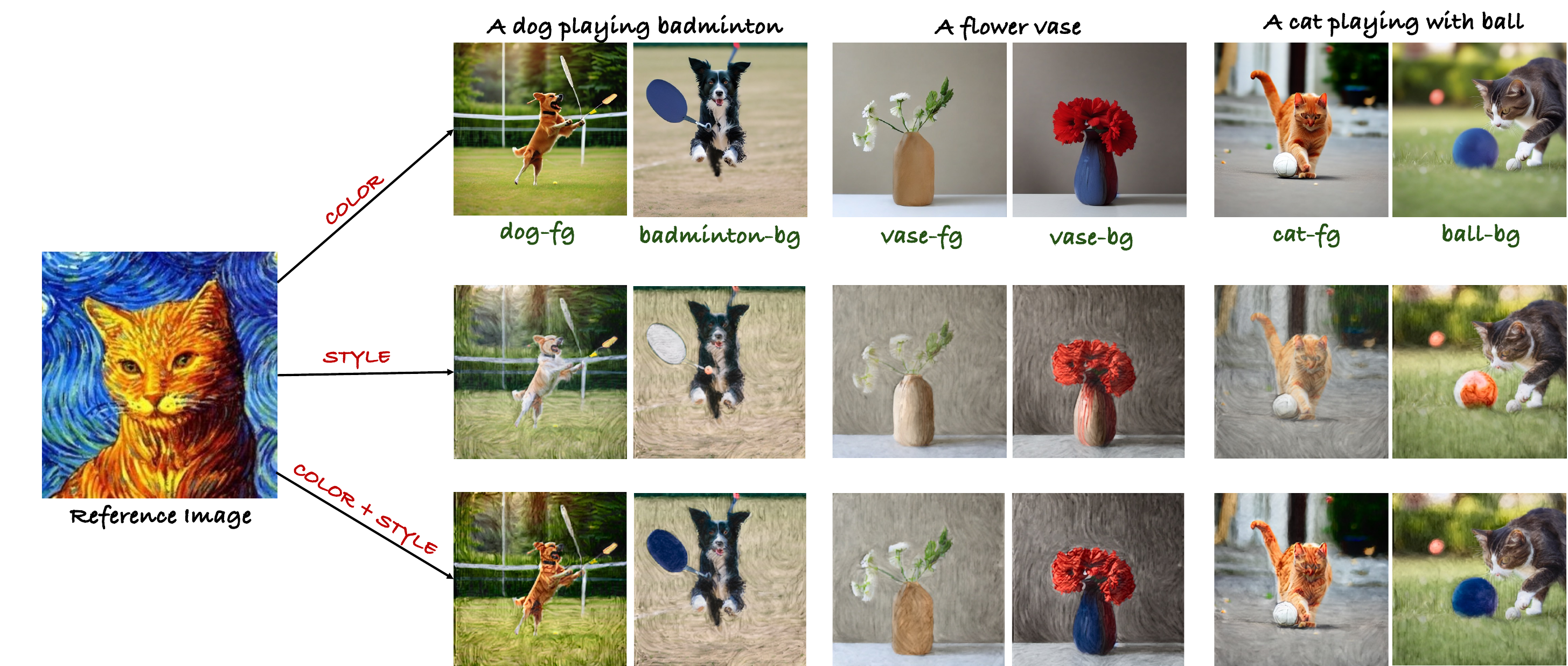}
  \caption{Additional qualitative results demonstrating disentangled conditioning on color and style attributes given a user-supplied reference image.}
  \label{fig:qual_supp}
\end{figure*}

We show additional qualitative results in Figure~\ref{fig:qual_supp} to further demonstrate the efficacy of our method in generating images conditioned on either of color or style attribute independently, or both color and style in a joint manner. In the first row, we show results for transferring colors from both background (bg) and foreground (fg) of the reference image to localised regions like the badminton, vase, and the ball in the generated images respectively. In the second row, we show results for global transfer of style from the reference image and as one can clearly note, the van gogh styled patterns are clearly visible in the generated images. Finally, in the third row, we generate images conditioned on both color and style attribute (e.g. see third column/last row where the ball has blue color transferred and the image follows van gogh style).

\subsection{Generating color variants}
\label{subsec:color_results}
\begin{figure*}[tb]
  \centering
  \includegraphics[width=\textwidth]{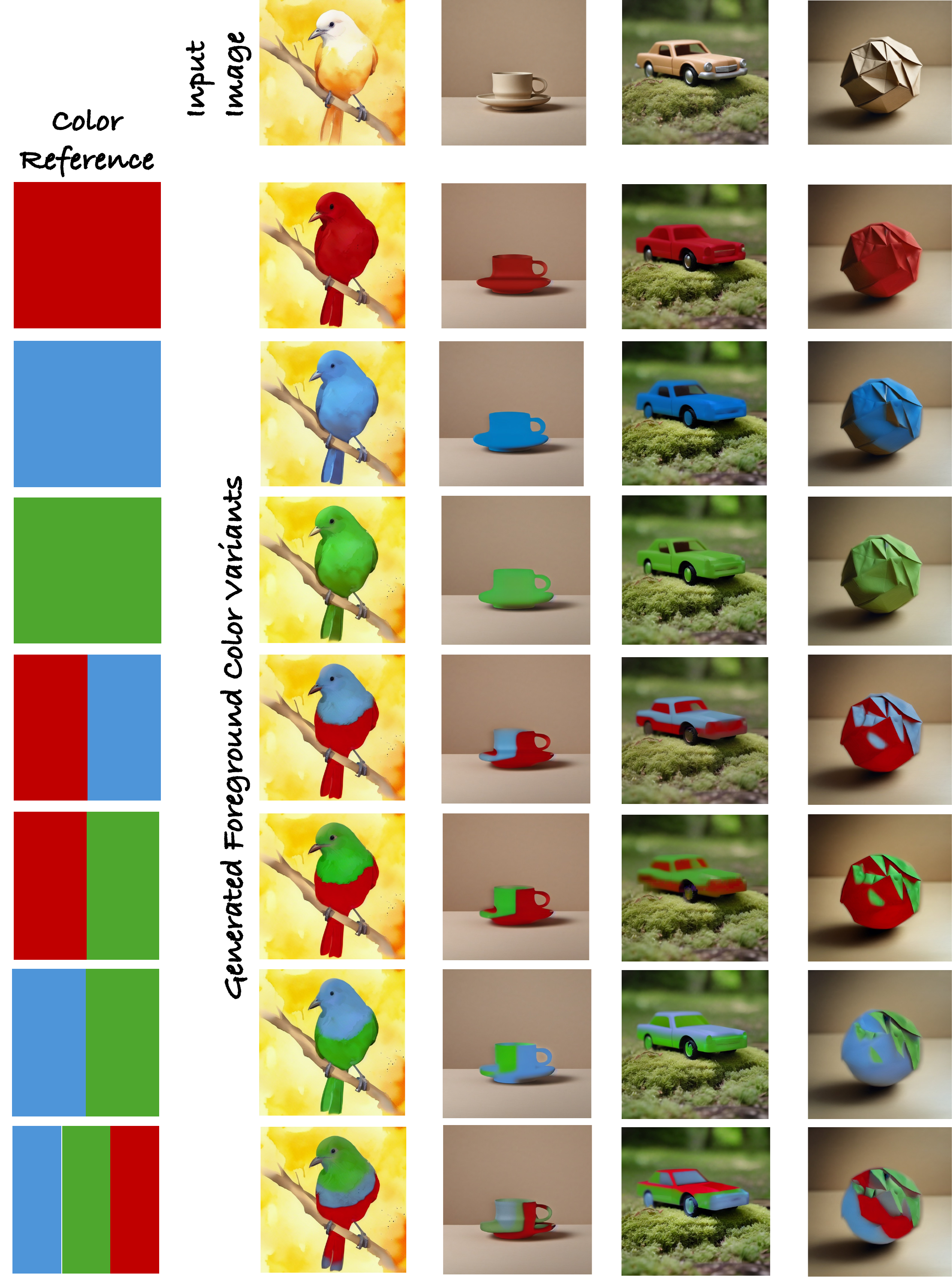}
  \caption{Color variants of image foreground.}
  \label{fig:color_variants}
\end{figure*}

We show additional color transfer results in Figure~\ref{fig:color_variants} while demonstrating its' utility in generating multiple color variants of a provided input image. We consider different colors (blue, green, red) and their combinations and show that our proposed method can effectively transfer color and all cases to the input image. For instance, note the first rows have the individual colors transferred for all of bird, cup toy car and ball. Similarly the rows 4-6 have combinations of two colors transferred, whereas the last row has all three colors (e.g. see first column where the generated bird has proportions of all three colors).

\subsection{Additional Style transfer results}
\label{subsec:style_results}

\begin{figure*}[tb]
  \centering
  \includegraphics[width=\textwidth]{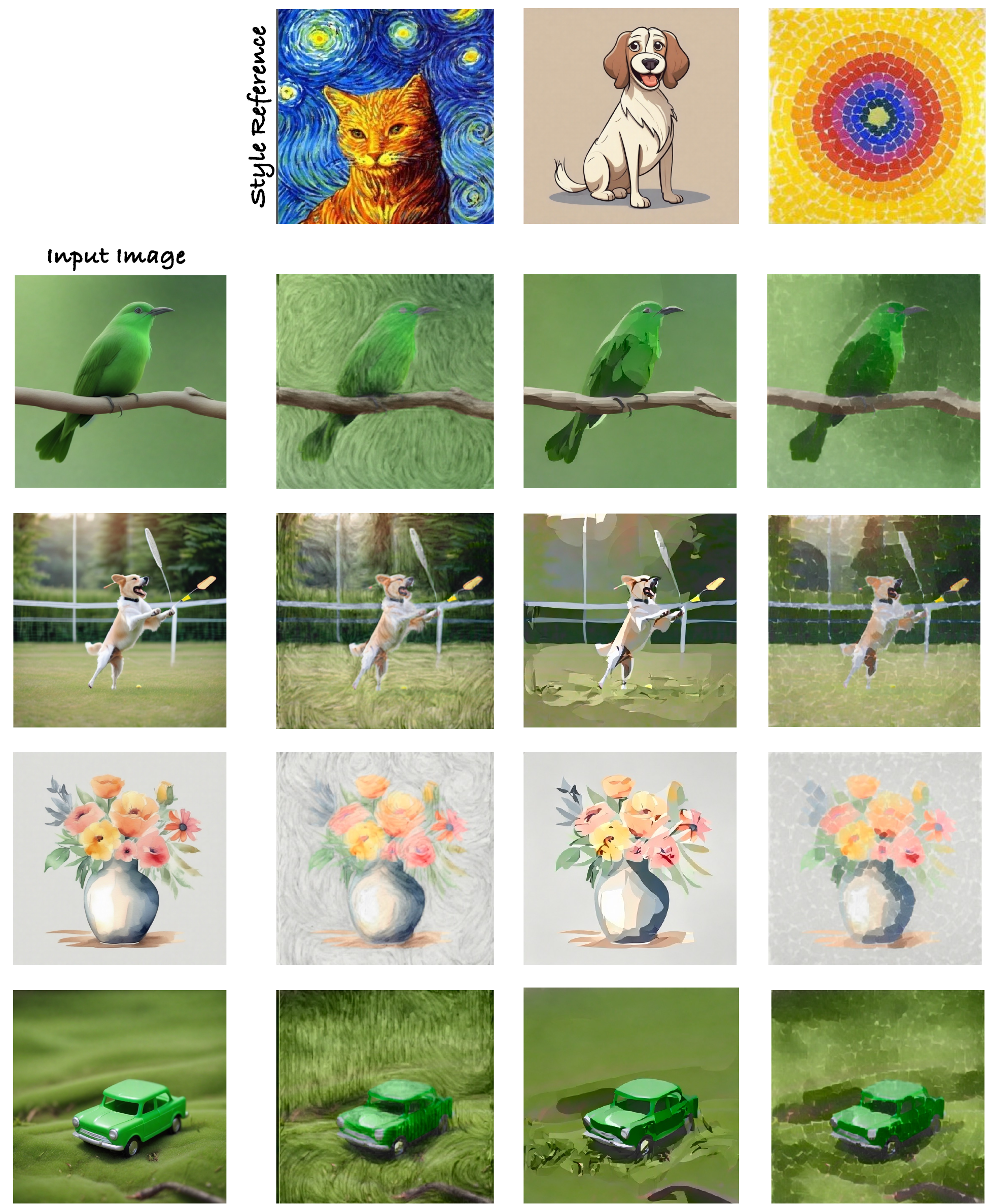}
  \caption{Additional style transfer results.}
  \label{fig:style_supp}
\end{figure*}

In Figure~\ref{fig:style_supp}, we show additional results to further demonstrate the effective of our proposed approach in transferring style from reference images in a disentangled manner without having an impact on the colors of the generated image. For instance, consider the last column where the reference image has a repeated squarebox type pattern. One can clearly note that all our generated images as well show the same pattern while still maintaining the colors from the original image.

\subsection{Limitations}
\label{subsec:limitations}
\begin{figure*}[tb]
  \centering
  \includegraphics[width=0.7\textwidth]{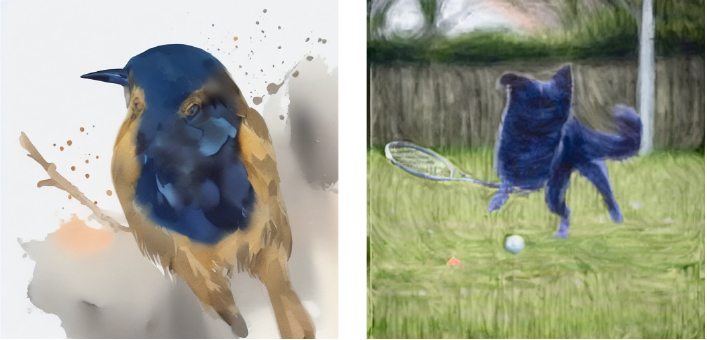}
  \caption{Limitations.}
  \label{fig:limitation}
\end{figure*}

In this Section, we briefly discuss a few limitations of the proposed approach. Firstly, as shown in Figure~\ref{fig:limitation}, the recoloring transforms in some cases lead to a loss of minute details e.g. bird's eye and details on the dog's face. Secondly, the masks obtained for the region of interest to be recolored are obtained using the cross-attention layers which, as shown in previous works \cite{agarwal2023star, chefer2023attend} as well, do not always accurately localize the region. In such a case, one can use high quality off-the-shelf methods \cite{kirillov2023segment, liu2023grounding} to obtain segmentation masks for improved localised color and style transfer. 

\end{appendices}

{\small
\bibliographystyle{ieee_fullname}
\bibliography{egbib}

\begin{thebibliography}{10}\itemsep=-1pt

\bibitem{agarwal2023star}
Aishwarya Agarwal, Srikrishna Karanam, KJ Joseph, Apoorv Saxena, Koustava Goswami, and Balaji~Vasan Srinivasan.
\newblock A-star: Test-time attention segregation and retention for text-to-image synthesis.
\newblock {\em Proceedings of the IEEE/CVF International Conference on Computer Vision}, 2023.

\bibitem{agarwal2023image}
Aishwarya Agarwal, Srikrishna Karanam, Tripti Shukla, and Balaji~Vasan Srinivasan.
\newblock An image is worth multiple words: Multi-attribute inversion for constrained text-to-image synthesis.
\newblock {\em arXiv preprint arXiv:2311.11919}, 2023.

\bibitem{alaluf2023cross}
Yuval Alaluf, Daniel Garibi, Or Patashnik, Hadar Averbuch-Elor, and Daniel Cohen-Or.
\newblock Cross-image attention for zero-shot appearance transfer.
\newblock {\em arXiv preprint arXiv:2311.03335}, 2023.

\bibitem{an2021artflow}
Jie An et~al.
\newblock Artflow: Unbiased image style transfer via reversible neural flows.
\newblock In {\em CVPR}, 2021.

\bibitem{chefer2023attend}
Hila Chefer, Yuval Alaluf, Yael Vinker, Lior Wolf, and Daniel Cohen-Or.
\newblock Attend-and-excite: Attention-based semantic guidance for text-to-image diffusion models.
\newblock {\em ACM Transactions on Graphics (TOG)}, 42(4):1--10, 2023.

\bibitem{chen2024anydoor}
Xi Chen, Lianghua Huang, Yu Liu, Yujun Shen, Deli Zhao, and Hengshuang Zhao.
\newblock Anydoor: Zero-shot object-level image customization.
\newblock In {\em Proceedings of the IEEE/CVF Conference on Computer Vision and Pattern Recognition}, pages 6593--6602, 2024.

\bibitem{chung2024style}
Jiwoo Chung, Sangeek Hyun, and Jae-Pil Heo.
\newblock Style injection in diffusion: A training-free approach for adapting large-scale diffusion models for style transfer.
\newblock In {\em Proceedings of the IEEE/CVF Conference on Computer Vision and Pattern Recognition}, pages 8795--8805, 2024.

\bibitem{deng2022stytr2}
Yingying Deng et~al.
\newblock Stytr2: Image style transfer with transformers.
\newblock In {\em CVPR}, 2022.

\bibitem{dhakar2015color}
Lokesh Dhakar.
\newblock Color thief.
\newblock {\em Retrieved}, 2015.

\bibitem{gal2022image}
Rinon Gal, Yuval Alaluf, Yuval Atzmon, Or Patashnik, Amit~H Bermano, Gal Chechik, and Daniel Cohen-Or.
\newblock An image is worth one word: Personalizing text-to-image generation using textual inversion.
\newblock {\em Proceedings of the International Conference on Learning Representations}, 2023.

\bibitem{gal2023encoder}
Rinon Gal, Moab Arar, Yuval Atzmon, Amit~H Bermano, Gal Chechik, and Daniel Cohen-Or.
\newblock Encoder-based domain tuning for fast personalization of text-to-image models.
\newblock {\em ACM Transactions on Graphics (TOG)}, 42(4):1--13, 2023.

\bibitem{gu2024mix}
Yuchao Gu, Xintao Wang, Jay~Zhangjie Wu, Yujun Shi, Yunpeng Chen, Zihan Fan, Wuyou Xiao, Rui Zhao, Shuning Chang, Weijia Wu, et~al.
\newblock Mix-of-show: Decentralized low-rank adaptation for multi-concept customization of diffusion models.
\newblock {\em Advances in Neural Information Processing Systems}, 36, 2024.

\bibitem{hertz2023style}
Amir Hertz, Andrey Voynov, Shlomi Fruchter, and Daniel Cohen-Or.
\newblock Style aligned image generation via shared attention.
\newblock {\em arXiv preprint arXiv:2312.02133}, 2023.

\bibitem{ho2020denoising}
Jonathan Ho, Ajay Jain, and Pieter Abbeel.
\newblock Denoising diffusion probabilistic models.
\newblock {\em Advances in neural information processing systems}, 33:6840--6851, 2020.

\bibitem{hong2023aespa}
Kibeom Hong et~al.
\newblock Aespa-net: Aesthetic pattern-aware style transfer networks.
\newblock In {\em ICCV}, 2023.

\bibitem{hossain2014whitening}
Maliha Hossain.
\newblock Whitening and coloring transformations for multivariate gaussian data.
\newblock {\em A slecture partly based on the ECE662 Spring}, 2014.

\bibitem{hu2021lora}
Edward~J Hu, Yelong Shen, Phillip Wallis, Zeyuan Allen-Zhu, Yuanzhi Li, Shean Wang, Lu Wang, and Weizhu Chen.
\newblock Lora: Low-rank adaptation of large language models.
\newblock {\em arXiv preprint arXiv:2106.09685}, 2021.

\bibitem{10.1145/3394171.3413853}
Zhiyuan Hu, Jia Jia, Bei Liu, Yaohua Bu, and Jianlong Fu.
\newblock Aesthetic-aware image style transfer.
\newblock In {\em Proceedings of the 28th ACM International Conference on Multimedia}, MM '20, page 3320–3329, New York, NY, USA, 2020. Association for Computing Machinery.

\bibitem{kirillov2023segment}
Alexander Kirillov, Eric Mintun, Nikhila Ravi, Hanzi Mao, Chloe Rolland, Laura Gustafson, Tete Xiao, Spencer Whitehead, Alexander~C Berg, Wan-Yen Lo, et~al.
\newblock Segment anything.
\newblock In {\em Proceedings of the IEEE/CVF International Conference on Computer Vision}, pages 4015--4026, 2023.

\bibitem{kumari2023multi}
Nupur Kumari, Bingliang Zhang, Richard Zhang, Eli Shechtman, and Jun-Yan Zhu.
\newblock Multi-concept customization of text-to-image diffusion.
\newblock In {\em Proceedings of the IEEE/CVF Conference on Computer Vision and Pattern Recognition}, pages 1931--1941, 2023.

\bibitem{li2024blip}
Dongxu Li, Junnan Li, and Steven Hoi.
\newblock Blip-diffusion: Pre-trained subject representation for controllable text-to-image generation and editing.
\newblock {\em Advances in Neural Information Processing Systems}, 36, 2024.

\bibitem{li2024diffstyler}
Shaoxu Li.
\newblock Diffstyler: Diffusion-based localized image style transfer.
\newblock {\em arXiv preprint arXiv:2403.18461}, 2024.

\bibitem{liu2023grounding}
Shilong Liu, Zhaoyang Zeng, Tianhe Ren, Feng Li, Hao Zhang, Jie Yang, Chunyuan Li, Jianwei Yang, Hang Su, Jun Zhu, et~al.
\newblock Grounding dino: Marrying dino with grounded pre-training for open-set object detection.
\newblock {\em arXiv preprint arXiv:2303.05499}, 2023.

\bibitem{mou2023t2i}
Chong Mou, Xintao Wang, Liangbin Xie, Jian Zhang, Zhongang Qi, Ying Shan, and Xiaohu Qie.
\newblock T2i-adapter: Learning adapters to dig out more controllable ability for text-to-image diffusion models.
\newblock {\em arXiv preprint arXiv:2302.08453}, 2023.

\bibitem{nichol2021glide}
Alex Nichol, Prafulla Dhariwal, Aditya Ramesh, Pranav Shyam, Pamela Mishkin, Bob McGrew, Ilya Sutskever, and Mark Chen.
\newblock Glide: Towards photorealistic image generation and editing with text-guided diffusion models.
\newblock {\em arXiv preprint arXiv:2112.10741}, 2021.

\bibitem{radford2021learning}
Alec Radford, Jong~Wook Kim, Chris Hallacy, Aditya Ramesh, Gabriel Goh, Sandhini Agarwal, Girish Sastry, Amanda Askell, Pamela Mishkin, Jack Clark, et~al.
\newblock Learning transferable visual models from natural language supervision.
\newblock In {\em International conference on machine learning}, pages 8748--8763. PMLR, 2021.

\bibitem{raffel2020exploring}
Colin Raffel, Noam Shazeer, Adam Roberts, Katherine Lee, Sharan Narang, Michael Matena, Yanqi Zhou, Wei Li, and Peter~J Liu.
\newblock Exploring the limits of transfer learning with a unified text-to-text transformer.
\newblock {\em The Journal of Machine Learning Research}, 21(1):5485--5551, 2020.

\bibitem{ramesh2022hierarchical}
Aditya Ramesh, Prafulla Dhariwal, Alex Nichol, Casey Chu, and Mark Chen.
\newblock Hierarchical text-conditional image generation with clip latents.
\newblock {\em arXiv preprint arXiv:2204.06125}, 2022.

\bibitem{rombach2022high}
Robin Rombach, Andreas Blattmann, Dominik Lorenz, Patrick Esser, and Bj{\"o}rn Ommer.
\newblock High-resolution image synthesis with latent diffusion models.
\newblock In {\em Proceedings of the IEEE/CVF Conference on Computer Vision and Pattern Recognition}, pages 10684--10695, 2022.

\bibitem{ruiz2023dreambooth}
Nataniel Ruiz, Yuanzhen Li, Varun Jampani, Yael Pritch, Michael Rubinstein, and Kfir Aberman.
\newblock Dreambooth: Fine tuning text-to-image diffusion models for subject-driven generation.
\newblock In {\em Proceedings of the IEEE/CVF Conference on Computer Vision and Pattern Recognition}, pages 22500--22510, 2023.

\bibitem{saharia2022photorealistic}
Chitwan Saharia, William Chan, Saurabh Saxena, Lala Li, Jay Whang, Emily Denton, Seyed Kamyar~Seyed Ghasemipour, Burcu~Karagol Ayan, S~Sara Mahdavi, Rapha~Gontijo Lopes, et~al.
\newblock Photorealistic text-to-image diffusion models with deep language understanding.
\newblock {\em arXiv preprint arXiv:2205.11487}, 2022.

\bibitem{shah2023ziplora}
Viraj Shah, Nataniel Ruiz, Forrester Cole, Erika Lu, Svetlana Lazebnik, Yuanzhen Li, and Varun Jampani.
\newblock Ziplora: Any subject in any style by effectively merging loras.
\newblock {\em arXiv preprint arXiv:2311.13600}, 2023.

\bibitem{shi2024instantbooth}
Jing Shi, Wei Xiong, Zhe Lin, and Hyun~Joon Jung.
\newblock Instantbooth: Personalized text-to-image generation without test-time finetuning.
\newblock In {\em Proceedings of the IEEE/CVF Conference on Computer Vision and Pattern Recognition}, pages 8543--8552, 2024.

\bibitem{sohl2015deep}
Jascha Sohl-Dickstein, Eric Weiss, Niru Maheswaranathan, and Surya Ganguli.
\newblock Deep unsupervised learning using nonequilibrium thermodynamics.
\newblock In {\em International conference on machine learning}, pages 2256--2265. PMLR, 2015.

\bibitem{sohn2023styledrop}
Kihyuk Sohn, Nataniel Ruiz, Kimin Lee, Daniel~Castro Chin, Irina Blok, Huiwen Chang, Jarred Barber, Lu Jiang, Glenn Entis, Yuanzhen Li, et~al.
\newblock Styledrop: Text-to-image generation in any style.
\newblock {\em arXiv preprint arXiv:2306.00983}, 2023.

\bibitem{song2020denoising}
Jiaming Song, Chenlin Meng, and Stefano Ermon.
\newblock Denoising diffusion implicit models.
\newblock In {\em ICLR}, 2021.

\bibitem{tewel2023key}
Yoad Tewel, Rinon Gal, Gal Chechik, and Yuval Atzmon.
\newblock Key-locked rank one editing for text-to-image personalization.
\newblock In {\em ACM SIGGRAPH 2023 Conference Proceedings}, pages 1--11, 2023.

\bibitem{tewel2024training}
Yoad Tewel, Omri Kaduri, Rinon Gal, Yoni Kasten, Lior Wolf, Gal Chechik, and Yuval Atzmon.
\newblock Training-free consistent text-to-image generation.
\newblock {\em arXiv preprint arXiv:2402.03286}, 2024.

\bibitem{voynov2023p+}
Andrey Voynov, Qinghao Chu, Daniel Cohen-Or, and Kfir Aberman.
\newblock $ p+ $: Extended textual conditioning in text-to-image generation.
\newblock {\em arXiv preprint arXiv:2303.09522}, 2023.

\bibitem{wang2024instantstyle}
Haofan Wang, Qixun Wang, Xu Bai, Zekui Qin, and Anthony Chen.
\newblock Instantstyle: Free lunch towards style-preserving in text-to-image generation.
\newblock {\em arXiv preprint arXiv:2404.02733}, 2024.

\bibitem{wang2024instantstyle2}
Haofan Wang, Peng Xing, Renyuan Huang, Hao Ai, Qixun Wang, and Xu Bai.
\newblock Instantstyle-plus: Style transfer with content-preserving in text-to-image generation.
\newblock {\em arXiv preprint arXiv:2407.00788}, 2024.

\bibitem{wang2023autostory}
Wen Wang, Canyu Zhao, Hao Chen, Zhekai Chen, Kecheng Zheng, and Chunhua Shen.
\newblock Autostory: Generating diverse storytelling images with minimal human effort.
\newblock {\em arXiv preprint arXiv:2311.11243}, 2023.

\bibitem{wei2023elite}
Yuxiang Wei, Yabo Zhang, Zhilong Ji, Jinfeng Bai, Lei Zhang, and Wangmeng Zuo.
\newblock Elite: Encoding visual concepts into textual embeddings for customized text-to-image generation.
\newblock In {\em Proceedings of the IEEE/CVF International Conference on Computer Vision}, pages 15943--15953, 2023.

\bibitem{xiao2023fastcomposer}
Guangxuan Xiao, Tianwei Yin, William~T Freeman, Fr{\'e}do Durand, and Song Han.
\newblock Fastcomposer: Tuning-free multi-subject image generation with localized attention.
\newblock {\em arXiv preprint arXiv:2305.10431}, 2023.

\bibitem{xu2024freetuner}
Youcan Xu, Zhen Wang, Jun Xiao, Wei Liu, and Long Chen.
\newblock Freetuner: Any subject in any style with training-free diffusion.
\newblock {\em arXiv preprint arXiv:2405.14201}, 2024.

\bibitem{yang2023paint}
Binxin Yang, Shuyang Gu, Bo Zhang, Ting Zhang, Xuejin Chen, Xiaoyan Sun, Dong Chen, and Fang Wen.
\newblock Paint by example: Exemplar-based image editing with diffusion models.
\newblock In {\em Proceedings of the IEEE/CVF Conference on Computer Vision and Pattern Recognition}, pages 18381--18391, 2023.

\bibitem{yang2024lora}
Yang Yang, Wen Wang, Liang Peng, Chaotian Song, Yao Chen, Hengjia Li, Xiaolong Yang, Qinglin Lu, Deng Cai, Boxi Wu, et~al.
\newblock Lora-composer: Leveraging low-rank adaptation for multi-concept customization in training-free diffusion models.
\newblock {\em arXiv preprint arXiv:2403.11627}, 2024.

\bibitem{ye2023ip}
Hu Ye, Jun Zhang, Sibo Liu, Xiao Han, and Wei Yang.
\newblock Ip-adapter: Text compatible image prompt adapter for text-to-image diffusion models.
\newblock {\em arXiv preprint arXiv:2308.06721}, 2023.

\bibitem{yu2023freedom}
Jiwen Yu, Yinhuai Wang, Chen Zhao, Bernard Ghanem, and Jian Zhang.
\newblock Freedom: Training-free energy-guided conditional diffusion model.
\newblock In {\em Proceedings of the IEEE/CVF International Conference on Computer Vision (ICCV)}, pages 23174--23184, October 2023.

\bibitem{zhang2023adding}
Lvmin Zhang, Anyi Rao, and Maneesh Agrawala.
\newblock Adding conditional control to text-to-image diffusion models.
\newblock In {\em ICCV}, 2023.

\bibitem{zhang2023prospect}
Yuxin Zhang, Weiming Dong, Fan Tang, Nisha Huang, Haibin Huang, Chongyang Ma, Tong-Yee Lee, Oliver Deussen, and Changsheng Xu.
\newblock Prospect: Expanded conditioning for the personalization of attribute-aware image generation.
\newblock {\em ACM Transactions on Graphics}, 42(6), dec 2023.

\bibitem{zhang2023inversion}
Yuxin Zhang, Nisha Huang, Fan Tang, Haibin Huang, Chongyang Ma, Weiming Dong, and Changsheng Xu.
\newblock Inversion-based style transfer with diffusion models.
\newblock In {\em Proceedings of the IEEE/CVF conference on computer vision and pattern recognition}, pages 10146--10156, 2023.

\bibitem{zhong2024multi}
Ming Zhong, Yelong Shen, Shuohang Wang, Yadong Lu, Yizhu Jiao, Siru Ouyang, Donghan Yu, Jiawei Han, and Weizhu Chen.
\newblock Multi-lora composition for image generation.
\newblock {\em arXiv preprint arXiv:2402.16843}, 2024.

\end{thebibliography}
}

\end{document}